\documentclass{ifacconf}

\usepackage{multirow}
\usepackage{graphicx}      
\usepackage{natbib}        
\begin{document}
\begin{frontmatter}

\title{Synthetic Data Augmentation Using GAN For Improved Automated Visual Inspection\thanksref{footnoteinfo}} 

\thanks[footnoteinfo]{This work was supported by the Slovenian Research Agency and the European Union’s Horizon 2020 program project STAR under grant agreement number H2020-956573.}

\author[1,2]{Jo\v{z}e M. Ro\v{z}anec}
\author[1,2]{Patrik Zajec} 
\author[3]{Spyros Theodoropoulos}
\author[4]{Erik Koehorst}
\author[5]{Bla\v{z} Fortuna}
\author[2]{Dunja Mladeni\'{c}}

\address[1]{Jo\v{z}ef Stefan Institute, Ljubljana, Slovenia, (e-mail: joze.rozanec@ijs.si, patrik.zajec@ijs.si)}
\address[2]{Jo\v{z}ef Stefan International Postgraduate School, Ljubljana, Slovenia, (e-mail: joze.rozanec@ijs.si, patrik.zajec@ijs.si)}
\address[3]{School of Electrical and Computer Engineering, National Technical University of Athens, Athens, Greece (e-mail: stheodoropoulos@mail.ntua.gr).}
\address[4]{Philips Consumer Lifestyle BV, Drachten, The Neatherlands (e-mail: erik.koehorst@philips.com)}
\address[5]{Qlector d.o.o., Ljubljana, Slovenia (e-mail: blaz.fortuna@qlector.com)}

\begin{abstract}                
Quality control is a crucial activity performed by manufacturing companies to ensure their products conform to the requirements and specifications. The introduction of artificial intelligence models enables to automate the visual quality inspection, speeding up the inspection process and ensuring all products are evaluated under the same criteria. In this research, we compare supervised and unsupervised defect detection techniques and explore data augmentation techniques to mitigate the data imbalance in the context of automated visual inspection. Furthermore, we use Generative Adversarial Networks for data augmentation to enhance the classifiers' discriminative performance. Our results show that state-of-the-art unsupervised defect detection does not match the performance of supervised models but can be used to reduce the labeling workload by more than 50\%. Furthermore, the best classification performance was achieved considering GAN-based data generation with AUC ROC scores equal to or higher than 0,9898, even when increasing the dataset imbalance by leaving only 25\% of the images denoting defective products. We performed the research with real-world data provided by \textit{Philips Consumer Lifestyle BV}.
\end{abstract}


\end{frontmatter}

\section{Introduction}\label{S:INTRODUCTION}


The increasing digitization of the manufacturing sector has enabled greater communication capabilities and tasks automation, producing an increasingly digital value chain \cite{benbarrad2021intelligent,papadopoulos2021towards}, which is one of the objectives of the Industry 4.0 paradigm. Quality control is considered one of the main areas where digitalization and the use of Artificial Intelligence provide new value to the manufacturing industry \cite{petrakieva2014http,chouchene2020artificial}, and has been applied in multiple scenarios \cite{beltran2020external,napoletano2021semi,obregon2021rule}. Artificial Intelligence promises to amend multiple issues associated with manual inspection, such as workers' fatigue, operator-to-operator inconsistency, and quality dependence on the employees' experience and well-being, among others \cite{see2012visual}. In addition, Artificial Intelligence enables great scalability by reducing manual work, increasing the speed of the inspection process, and allowing the execution of the inspection process in a continuous manner \cite{garvey2018framework,escobar2018machine,chouchene2020artificial}. The resulting increase in product quality directly impacts the whole production chain and the business, allowing to trace defect root causes, reducing rework in the production process, and avoiding costly disruptions in the supply chain.

Among the challenges posed by the development of Artificial Intelligence models for visual inspection, we find data gathering \cite{ren2021state}, data labeling (in the case of supervised models), increased class imbalance over time (according to the increasing products' quality), and models' explainability \cite{meister2021explainability}.

We frame the visual inspection problem as a supervised learning problem. This paper addresses the problems of class imbalance and data scarcity regarding defective pieces. These are critical visual inspection problems in manufacturing since increasing products' quality only increases the class imbalance over time and makes it harder to improve the models' classification performance. It is thus imperative to devise strategies that can be used to mitigate such an issue. To that end, we conduct a series of experiments to (a) compare our classification models to a State-Of-The-Art (SOTA) unsupervised defect detection method, (b) understand how using trivial techniques to balance the dataset influences the classifier's performance, (c) we devote particular attention to the use of Generative Adversarial Network (GAN) models to generate synthetic images of defective parts and evaluate how such a balanced dataset influences the classifier's discrimination performance, and (d) simulate more severe class imbalances, to understand how would the classifiers could behave in the future, as we expect a higher scarcity of defective parts when products' quality increases. We developed the machine learning models with images provided by the \textit{Philips Consumer Lifestyle BV} corporation. The dataset comprises shaver images classified into three categories based on the defects observed regarding logo printing: good, double print, and interrupted print.

The rest of this paper is structured as follows: Section~\ref{S:RELATED-WORK} presents related work,  Section~\ref{S:USE-CASE} describes the use case on which we conducted the research, Section~\ref{S:EXPERIMENTS} describes the methodology we followed, and the experiments we performed, and Section~\ref{S:RESULTS} presents the results we obtained, and their implications. Finally, in Section~\ref{S:CONCLUSION}, we provide our conclusions and outline future work.

\section{Related Work}\label{S:RELATED-WORK}
Automated visual inspection is an automatic form of non-destructive testing in quality control, achieved through one or multiple cameras that provide visual input and visual processing techniques that allow for detecting defects (\cite{czimmermann2020visual}). The visual inspection aims to identify functional and cosmetic defects (\cite{chin1982automated}). Automated visual inspection techniques can be either supervised (requiring labeled data) or unsupervised (requiring no labeling). SOTA image processing techniques involve the use of deep learning (\cite{pouyanfar2018survey}). Authors frequently use pre-trained models, either as feature extractors (to avoid costly feature engineering) or for end-to-end learning, either by training the model from scratch or fine-tuning a pre-trained one (\cite{wang2018fast}).

The decreased cost of sensors and their implementation in production lines in the context of Industry 4.0 produces greater amounts of digital data (\cite{benbarrad2021intelligent}), which can be leveraged for automated visual inspection (\cite{peres2020industrial}). Given that the reliability of manufactured products is of utmost importance, inspecting all parts is desired. Such inspection translates into great costs, which can be significantly decreased by automating the visual inspection process. The main benefits of such automation are the savings related to human labor costs, the ability to match production and inspection speeds, and error traceability, which provides relevant information to guide management decisions (\cite{chin1982automated}).

Successful automated visual inspection implementations report using Artificial Intelligence models to ensure the quality of printing (\cite{villalba2019deep}), steel surface defects (\cite{jia2004intelligent}), or assess the quality of manufactured vehicle parts (\cite{chouchene2020artificial}). The most commonly used algorithms are the Support Vector Machines, k-Nearest Neighbours, Multi-Layer Perceptrons (MLPs), and Convolutional Neural Networks (CNNs) (\cite{ren2021state}).

Labeling data availability is critical to train and to enhance supervised artificial intelligence models. While the acquisition of data can be cheap, the labeling process is expensive in terms of time and costs since it requires human persons to review the images and provide labels for them. In addition, when inspecting data related to products' quality, it is to expect a strong class imbalance: most of the data instances correspond to good products, and just a minor proportion of them refer to defective pieces. If the class imbalance is not reduced, (a) most of the labeling time will be devoted to images of non-defective products, which provide little information to the model, (b) and operators will be  prone to skip cases that correspond to defective pieces, confusing such an instance with the many ones that correspond to non-defective products. Given the products' quality improves over time, such class imbalance will be more pronounced. An alternative to expensive data labeling could be the generation of synthetic images corresponding to defective parts. Such a solution would require an initial dataset but enable an unlimited supply of synthetic images, which could be used to enrich the dataset and eventually avoid most of the labeling procedure. Furthermore, such an approach could also be effective in settings where different cameras are installed on each production line, thus producing images with different characteristics (e.g., different color scales). Such a setting could likely prevent using the images of different production lines in a single dataset but require different datasets for images of different characteristics. Using synthetic images could alleviate much of the manual workload required to gather the data over time.

Datasets built for the supervised defect detection use case are usually small and imbalanced, given the cost required to obtain and label data instances from a dataset. It is thus important to choose such data augmentation methods that can deliver value regardless of the size of the dataset. While classical data augmentation techniques, such as SMOTE, ADASYN, or more recent variants, such as DeepSMOTE are frequently used, GANs can learn a function that can approximate model distribution close to a true distribution, generating high fidelity (\cite{gan_imbalance_survey}). While the dataset size could prevent training autoencoders or GANs from scratch, transfer learning can leverage pre-trained neural networks and fine-tune them to comply with the task. There has been extensive research on applying transfer learning to GANs, but many methods relying on traditional weight fine-tuning still fall short when faced with small datasets (\cite{Wang2018TransferringGG}). A solution successfully applied to BigGAN and SN-GAN is to focus only on fine-tuning the batch normalization layers, thus reducing the number of adjustable weights (\cite{Noguchi2019ImageGF}). The latter is intuitively equivalent to selecting those features relevant to the current dataset. Some preliminary results using the above method on a pre-trained BigGAN, have shown a significant increase in classifier accuracy after augmenting the original dataset. While imperfect, the images produced seem to retain some of the features needed for classification. Combining generated and original images to increase quality is possible, as done by \cite{metasatoshi19}.

GANs have been successfully used in various industrial scenarios to tackle class imbalance in defect detection problems. For example, \cite{Jain2020SyntheticDA} compare three GANs (DC-GAN, Auxiliary Classifier GAN (AC-GAN), and Information-theoretic GAN (InfoGAN)) applied for steel streep defect detection. In their research, GANs enable an improved classification sensitivity from the baseline (consisting only of graphical image transformations), with DC-GAN performing the best. DC-GAN was combined with geometric transformations in a fiber layup inspection setting, providing high-quality synthetic images for datasets with less than fifty representative original images per class \cite{Meister2021Synth}. \cite{Wang_2021}, on the other hand, applied DC-GANs to generate images regarding solar cell defects, achieving substantial improvements only after combining real and synthetic images using a randomized boolean mosaic approach.

This research uses classical data augmentation techniques and the Lightweight GAN (\cite{liu2020towards}) to augment the dataset with synthetic data instances. Classical data augmentation techniques are applied to the feature vectors that result from the feature extraction process. At the same time, the Lightweight GAN is used to create synthetic images, which are added to the regular dataset and then processed to extract the required features. 

\section{Use Case}\label{S:USE-CASE}
This research considered the visual inspection task performed on shavers manufactured by \textit{Philips Consumer Lifestyle BV}. This non-destructive quality control testing focuses on a cosmetic aspect of the product quality: whether the \textit{Philips Consumer Lifestyle BV} company logo was printed correctly and eventually detects defective printings and determines the type of defective printing. To perform such an inspection, the company uses different pad-printing setups. Countless products are manufactured daily, and they are handled and inspected. Defective products are removed from the production line. An automated visual quality inspection system could strongly reduce manual work and decrease the inspection time by up to 40\%. 

For this research, a labeled dataset of 3.518 images was provided. The effort required to label such a dataset is close to eight man-hours and could rise to twenty-four man-hours if some triaging labeling strategy is introduced to increase the quality of the labeling. Images were labeled into three categories: "good" (no defect was detected), "double printing", and "interrupted printing". We tackle the problem as a binary classification and a multiclass classification task.

\section{Methodology and Experiments}\label{S:EXPERIMENTS}

We framed the automated defect detection as a binary classification problem (defective vs. non-defective) and a multiclass classification problem (classifying the images into one of three possible categories: good, double print, or interrupted print). To measure the models' discriminative power, we computed the AUC ROC metric. In the case of multiclass classification, we adopted the "one-vs-rest" heuristic. The heuristic splits the dataset to compute the AUC ROC metric for each class. Then a final value is obtained by calculating the weighted average, weighting the AUC ROC scores of each class by the proportion of true instances of that class in the dataset.

To evaluate the models, we divided the dataset using a stratified ten-fold cross-validation (\cite{zeng2000distribution,kuhn2013applied}). When performing data augmentation to balance the dataset, we considered the instances in the training set of the current k-fold cross-validation instance. While this meant computing ten times the instances for data augmentation, where necessary, at the same time ensured the instances were as close as possible to the data distribution observed in the training set when training the model. We used the Lightweight GAN to perform GAN-based data augmentation. To assess the quality of the synthetic images when compared to the original ones, we measured the Fr\'echet Inception Distance, which is defined as the squared Wasserstein metric between two multidimensional Gaussian distributions: the distribution that corresponds to features from the real-world images, and the one that corresponds to features from the synthetic images.

We used a pre-trained ResNet-18 model to extract average pool layer features. This allowed us to obtain a 512 values long vector per image, which we then reduced to a final feature vector of \textit{K} features considering $K=\sqrt{N}$, with N equal to the number of data instances in the train set. Finally, feature selection was made by selecting the top \textit{K} features when ranking them by their mutual information score.

Throughout the experiments, we used a Multi-layer Perceptron classifier \cite{Rosenblatt1958ThePA} as a baseline and compared it in different settings and against other models. All experiments were performed with four variations of the dataset: (a) considering all data available and reducing the number of images corresponding to defective parts so that only (b) 75\%, (c) 50\%, and (d) 25\% are left. Doing so enabled us to understand how the models perform under the existing circumstances and how their performance can change over time when the products' quality improves, and the proportion of defective products decreases. Furthermore, a good classifier's performance under heavier imbalance would also inform data collection and labeling efforts. For example, the need for fewer defective product samples could shorten the data collection time and reduce the number of images to label, with the consequent resource gains (time, people availability, and money).

In the following subsections, we describe the experiments performed and their rationale. We made the code available in a publicly accessible repository to promote research reproducibility \footnote{The repository URL will be provided upon paper acceptance. The dataset will remain confidential, as requested by \textit{Philips Consumer Lifestyle BV}.}.

\subsection{Experiment 1: compare supervised and SOTA unsupervised models}
The first experiment we performed was to compare three models: the baseline supervised model (MLP), a Gradient Boosted Tree classifier \footnote{For this research, we use the Catboost implementation.} (which are frequently cited in the literature (\cite{yorulmucs2021predictive,vzidek2016diagnostics})), and DRAEM (\cite{zavrtanik2021draem}), which is considered a SOTA model for unsupervised anomaly detection. DRAEM trains an autoencoder to reconstruct images to look like non-defective pieces. The original and reconstructed images are then fed to a discriminative sub-network, which identifies the anomalous regions to create an anomaly map. However, given that this model can only inform if a defect is present, it cannot be compared against a supervised model in a multiclass setting. We used the following model parameters: for the baseline model, the MLP was built with two dense layers (with 512 and 100 features), with an intermediate ReLU activation between both dense layers and a softmax activation at the output. The GBT model was instantiated with a maximum tree depth of ten, the multiclass loss function, and trained over sixty iterations.

This experiment aimed to determine whether and how much better the supervised models are compared to unsupervised SOTA models. 

\subsection{Experiment 2: effect of classic oversampling on supervised models}
After validating that the supervised models outperformed the unsupervised SOTA model (the results are presented in Table \ref{T:RESULTS-ROC} in Section \ref{S:RESULTS}), we explored if their performance could be enhanced by enlarging the dataset with synthetic data following three classic oversampling strategies: RANDOM (oversampling the minority classes by picking samples at random with replacement), ADASYN, and SMOTE.

\subsection{Experiment 3: effect of GAN-based oversampling on supervised models}
As described in Section \ref{S:RELATED-WORK}, a growing body of research is focused on leveraging GANs to create synthetic data to mitigate data scarcity. A strong advantage of the GANs is that they learn to produce realistic data instances based on the feedback provided by a discriminator, which attempts to identify which images were artificially generated and which ones correspond to the original dataset.

After corroborating the positive outcome of oversampling on the supervised models, we explored if synthetic images generated with GANs could further enhance the classifiers' performance. Our premise was that the similarity between the original and synthetic image distributions and the greater amount of data would boost the classifiers' performance. We thus explore whether the inclusion of synthetic images generated with a Lightweight GAN model enhances the models' performance and how additional class imbalance affects the GANs and the classifiers' performance. Given the poor performance of the Gradient Boosted Trees model, we experimented with the baseline model only.

We trained the Lightweight GAN models \footnote{We trained the Lightweight GAN based on the implementation available in the following repository: (the repository will be provided upon paper acceptance).} on a Tesla V100 32GB GPU, considering 20.000 iterations, investing nearly two hours per model trained. While we also explored other GANs, such as the StyleGAN, we found that they required many more resources without achieving a higher image quality. E.g., training a StyleGAN took us nearly six times the time required to train a Lightweight GAN model.

\subsection{Experiment 4: enhancing classification performance with a custom Neural Network model}
Finally, given the excellent results obtained in the experiments above with the baseline MLP model, we explored multiple neural network architectures to find some simple architectures that could achieve a better classification performance. We assumed that while using embeddings from a pre-trained ResNet18 model provided excellent features for a classifier, better features could be obtained by a simple CNN model trained directly on our dataset, given the visual simplicity of the images at hand. Furthermore, we trained the model on the original dataset (without data oversampling), considering two cases: training with an unweighted and a weighted loss. The weighted loss is a class imbalance mitigation technique that allows weighting the loss differently for different samples, considering whether they belong to the majority or a minority class. These two settings make the CNN comparable to the supervised models in \textit{Experiment 1} (for the unweighted loss), and \textit{Experiment 2} and \textit{Experiment 3} when using the weighted loss.

The final architecture was trained using the Adam optimizer \cite{kingma2014adam}, the categorical cross-entropy, and using softmax as the output activation function.

\begin{table*}[ht!]
\begin{tabular}{|c|l|rrrr|rrrr|}
\hline
\multirow{2}{*}{\textbf{Experiment}} & \multicolumn{1}{c|}{\multirow{2}{*}{\textbf{Model}}} & \multicolumn{4}{c|}{\textbf{ROC   AUC (binary)}} & \multicolumn{4}{c|}{\textbf{ROC   AUC (multiclass)}} \\ \cline{3-10} 
 & \multicolumn{1}{c|}{} & \multicolumn{1}{l|}{\textbf{100\%}} & \multicolumn{1}{l|}{\textbf{75\%}} & \multicolumn{1}{l|}{\textbf{50\%}} & \multicolumn{1}{l|}{\textbf{25\%}} & \multicolumn{1}{l|}{\textbf{100\%}} & \multicolumn{1}{l|}{\textbf{75\%}} & \multicolumn{1}{l|}{\textbf{50\%}} & \multicolumn{1}{l|}{\textbf{25\%}} \\ \hline
\multirow{3}{*}{\textbf{1}} & \textbf{Baseline} & \multicolumn{1}{r|}{0,9894} & \multicolumn{1}{r|}{0,9889} & \multicolumn{1}{r|}{0,9880} & 0,9835 & \multicolumn{1}{r|}{0,9897} & \multicolumn{1}{r|}{0,9882} & \multicolumn{1}{r|}{0,9881} & 0,9837 \\ \cline{2-10} 
 & \textbf{GBT} & \multicolumn{1}{r|}{0,9697} & \multicolumn{1}{r|}{0,9703} & \multicolumn{1}{r|}{0,9634} & 0,9516 & \multicolumn{1}{r|}{0,9696} & \multicolumn{1}{r|}{0,9698} & \multicolumn{1}{r|}{0,9619} & 0,9470 \\ \cline{2-10} 
 & \textbf{DRAEM} & \multicolumn{1}{r|}{0,8624} & \multicolumn{1}{r|}{0,8624} & \multicolumn{1}{r|}{0,8624} & 0,8624 & \multicolumn{1}{r|}{-} & \multicolumn{1}{r|}{-} & \multicolumn{1}{r|}{-} & - \\ \hline
\multirow{6}{*}{\textbf{2}} & \textbf{OS - RANDOM(baseline)} & \multicolumn{1}{r|}{0,9910} & \multicolumn{1}{r|}{0,9903} & \multicolumn{1}{r|}{0,9903} & 0,9850 & \multicolumn{1}{r|}{0,9912} & \multicolumn{1}{r|}{0,9905} & \multicolumn{1}{r|}{\textit{0,9905}} & 0,9850 \\ \cline{2-10} 
 & \textbf{OS - RANDOM(GBT)} & \multicolumn{1}{r|}{0,9728} & \multicolumn{1}{r|}{0,9710} & \multicolumn{1}{r|}{0,9650} & 0,9512 & \multicolumn{1}{r|}{0,9732} & \multicolumn{1}{r|}{0,9714} & \multicolumn{1}{r|}{0,9651} & 0,9505 \\ \cline{2-10} 
 & \textbf{OS - ADASYN(baseline)} & \multicolumn{1}{r|}{0,9910} & \multicolumn{1}{r|}{0,9906} & \multicolumn{1}{r|}{0,9899} & 0,9853 & \multicolumn{1}{r|}{0,9912} & \multicolumn{1}{r|}{0,9907} & \multicolumn{1}{r|}{0,9900} & \textit{0,9851} \\ \cline{2-10} 
 & \textbf{OS - ADASYN(GBT)} & \multicolumn{1}{r|}{0,9760} & \multicolumn{1}{r|}{0,9727} & \multicolumn{1}{r|}{0,9693} & 0,9575 & \multicolumn{1}{r|}{0,9762} & \multicolumn{1}{r|}{0,9731} & \multicolumn{1}{r|}{0,9693} & 0,9572 \\ \cline{2-10} 
 & \textbf{OS - SMOTE(baseline)} & \multicolumn{1}{r|}{0,9902} & \multicolumn{1}{r|}{0,9910} & \multicolumn{1}{r|}{0,9899} & 0,9854 & \multicolumn{1}{r|}{0,9903} & \multicolumn{1}{r|}{0,9911} & \multicolumn{1}{r|}{0,9900} & 0,9847 \\ \cline{2-10} 
 & \textbf{OS - SMOTE(GBT)} & \multicolumn{1}{r|}{0,9727} & \multicolumn{1}{r|}{0,9686} & \multicolumn{1}{r|}{0,9645} & 0,9551 & \multicolumn{1}{r|}{0,9726} & \multicolumn{1}{r|}{0,9684} & \multicolumn{1}{r|}{0,9640} & 0,9534 \\ \hline
\textbf{3} & \textbf{OS - GAN} & \multicolumn{1}{r|}{\textbf{0,9965}} & \multicolumn{1}{r|}{\textbf{0,9954}} & \multicolumn{1}{r|}{\textbf{0,9948}} & \textbf{0,9898} & \multicolumn{1}{r|}{\textbf{0,9966}} & \multicolumn{1}{r|}{\textbf{0,9955}} & \multicolumn{1}{r|}{\textbf{0,9955}} & \textbf{0,9899} \\ \hline
\multirow{2}{*}{\textbf{4}} & \textbf{CNN} & \multicolumn{1}{r|}{\textit{0,9934}} & \multicolumn{1}{r|}{\textit{0,9931}} & \multicolumn{1}{r|}{\textit{0,9907}} & \textit{0,9859} & \multicolumn{1}{r|}{\textit{0,9928}} & \multicolumn{1}{r|}{\textit{0,9918}} & \multicolumn{1}{r|}{0,9900} & \textit{0,9851} \\ \cline{2-10} 
 & \textbf{CNN + loss weighting} & \multicolumn{1}{r|}{0,9919} & \multicolumn{1}{r|}{0,9896} & \multicolumn{1}{r|}{0,9896} & 0,9854 & \multicolumn{1}{r|}{0,9916} & \multicolumn{1}{r|}{0,9892} & \multicolumn{1}{r|}{0,9892} & 0,9845 \\ \hline
\end{tabular}
\caption{ROC AUC (binary and multiclass) obtained when conducting the experiments. 100\%, 75\%, 50\%, and 25\% denote the percentage of images with defective parts from the original training set, effectively left in the training set, to simulate higher class imbalances. The best results are bolded, second-best results are highlighted in italics.}
\label{T:RESULTS-ROC}
\end{table*}

\section{Results}\label{S:RESULTS}
We present the results regarding the classifiers' discriminative power measured with the AUC ROC in Table \ref{T:RESULTS-ROC}. In addition, we detail the Fr\'echet Inception Distance obtained when generating images with the Light-weight GAN in Table \ref{T:RESULTS-FID}. When creating synthetic instances with a GAN, we found that the best quality was achieved for the images that correspond to "interrupted prints". In contrast, surprisingly, the worst quality was achieved for the "good" images. Furthermore, we observed that the classifier frequently confused "good" and "interrupted print" images. After carefully analyzing those instances, we concluded that errors were likely made when labeling data instances to both categories, though the errors were not confirmed. Nevertheless, given how hard it is to identify interrupted prints in some instances, we consider this a reasonable explanation.

Our main findings regarding the classification models are that (a) supervised models have a very good performance and strongly surpass the unsupervised models; (b) data augmentation techniques always improve the discriminative power of the classifier in binary and multiclass settings. Nevertheless, (c) the CNN model achieved the second-best results. We detail the experiment results and their implications in the following subsections.

\begin{table}
\centering
\begin{tabular}{l|c|c|c}
MODEL & Good & Double print & Interrupted print \\ \hline
Light-weight GAN & 40.80 & 39.49 & 29.68 \\ 
\end{tabular}
\caption{Fr\'echet Inception Distance score for each class.}
\label{T:RESULTS-FID}
\end{table}

\subsection{Experiment 1: compare supervised and SOTA unsupervised models}
When comparing the supervised models against DRAEM, we found that the supervised models always surpassed DRAEM in the binary classification setting. In the worst case, they outperformed DRAEM by 0,0892 when measuring AUC ROC. The baseline model outperformed the GBT in all cases among the supervised models. Since DRAEM does not output specific classes, it was excluded from the comparison when computing ROC AUC in the multiclass setting. 

Given the results presented above, we concluded that while unsupervised models have the advantage of working on unlabeled data, supervised models can be preferred to ensure better quality inspection. Nevertheless, unsupervised defect detection methods can be valuable for the labeling procedure. They can filter images and prioritize those that contain defects over those that do not. Such an approach could automate the labeling of good images (saving more than 50\% of manual work) and let the labeling people concentrate on the defective parts. Furthermore, following \cite{lehr2020automated}, means to cluster defective parts can be explored, to automate further or ease manual labeling of images concerning defective parts.

\subsection{Experiment 2: effect of classic oversampling on supervised models}
When applying RANDOM, ADASYN, and SMOTE oversampling techniques, we observed that the supervised models almost always improved their performance. However, three exceptions were found, all of them for the GBT model: (a) for binary ROC AUC, when using SMOTE and preserving 75\% of images regarding defective parts, (b) for binary ROC AUC, when using RANDOM oversampling and preserving 25\% of images regarding defective parts, and (c) for multiclass ROC AUC, when using SMOTE and preserving 75\% of images regarding defective parts. Our baseline model outperformed the GBT model in all cases, which led to the decision to continue the rest of the experiments considering only the baseline model. 

\subsection{Experiment 3: effect of GAN-based oversampling on supervised models}
Given the positive outcomes in \textit{Experiment 2}, we explored the effect of GAN-oversampling on the baseline model. We found that GAN oversampling led to the best ROC AUC results, without exception, with metric values ranging between 0,9898 and 0,9966. In addition, GAN oversampling enabled the model to improve between 0,0063 and 0,0074 points compared to the same model trained without oversampling.

\subsection{Experiment 4: enhancing classification performance with a custom Neural Network model}
When comparing the baseline MLP to the CNN model in the two settings (weighted and unweighted loss), we found that the CNN model trained with the unweighted loss achieved the second-best performance in all cases. However, contrary to the performance increase observed in Experiments 2 and 3 with data augmentation, loss weighting negatively affected the model. Though it achieved better results than models in Experiment 2 for the original dataset, it could not match their performance when a stronger imbalance was introduced by preserving 75\%, 50\%, or 25\% of the images of defective pieces. 

\section{Conclusions}\label{S:CONCLUSION}
This research compared supervised and unsupervised classification models and explored data augmentation techniques to avoid class imbalance in the automated visual inspection use case. We confirmed that supervised models outperform SOTA unsupervised defect detection models. Furthermore, the best classifier discrimination performance was achieved with GAN-based data augmentation and an MLP classifier, using a pre-trained ResNet18 as a feature extractor. We used the Lightweight GAN implementation to generate images regarding defective pieces. The GAN achieved good quality images while consuming few resources. From the results obtained, we consider (a) unsupervised models can be of great utility to identify images corresponding to non-defective pieces and automatically label them, thus reducing labeling work for at least 50\%; (b) GANs can be used to reduce further the effort required to gather (e.g., by reducing the waiting time to get a defective piece) and label images of defective pieces, by generating synthetic images; and (c) supervised classification models achieve the best performance, and thus are the best choice when deploying them to production, to automate the visual inspection. Future work will focus on designing a pipeline that implements (a) and (b), along with explainable artificial intelligence techniques, to enhance labeling and manual revision. In particular, the pipeline will reduce the workload, provide hints to the users to ease the labeling effort, and monitor users' attention to ensure the best labeling quality.

\begin{ack}
The authors acknowledge the valuable input and help Yvo van Vegten from \textit{Philips Consumer Lifestyle BV}
\end{ack}

\bibliography{main}             
                                                   






\end{document}